\documentclass[conference]{IEEEtran}
\IEEEoverridecommandlockouts
\usepackage{cite}
\usepackage{amsmath,amssymb,amsfonts}
\usepackage{algorithmic}
\usepackage{graphicx}
\usepackage{textcomp}
\usepackage{xcolor}
\def\BibTeX{{\rm B\kern-.05em{\sc i\kern-.025em b}\kern-.08em
    T\kern-.1667em\lower.7ex\hbox{E}\kern-.125emX}}

\usepackage{subfigure}
\usepackage{algorithm,algorithmic}

\usepackage{tabularx,ragged2e}

\def\BibTeX{{\rm B\kern-.05em{\sc i\kern-.025em b}\kern-.08em
    T\kern-.1667em\lower.7ex\hbox{E}\kern-.125emX}}

\newcolumntype{R}{>{\raggedleft\arraybackslash}X}
\newcolumntype{L}{>{\raggedright\arraybackslash}X}

\newif\ifdraft\drafttrue

\ifdraft

\fi

\usepackage{balance}
\usepackage{lastpage}

\usepackage{fancyhdr}
\begin{document}

\title{\huge Bayesian Optimization for Developmental Robotics with Meta-Learning by Parameters Bounds Reduction\\
\thanks{This work was supported by the EU FEDER funding through the FUI PIKAFLEX project and by the French National Research Agency (ANR), through the ARES labcom project under grant ANR 16-LCV2-0012-01, and by the CHIST-ERA EU project "Learn-Real"}% <-this % stops a space
}

%\thanks{$^{1}$ LIRIS, CNRS UMR 5205, Ecole Centrale de Lyon, France
%         {\tt\small name.surname@ec-lyon.fr}}%
% \thanks{This work is supported by the EU FEDER funding (FUI PIKAFLEX project) and by the French National Research Agency, l'Agence Nationale de Recherche (ARES LabCom under grant ANR 16-LCV2-0012-01).}% <-this % stops a space
% }

% City, Country}
% \and
% City, Country}
% \and
% City, Country}
% }

\author{\IEEEauthorblockN{1\textsuperscript{st} Maxime Petit}
\IEEEauthorblockA{\textit{LIRIS, CNRS UMR 5205} \\
\textit{Ecole Centrale de Lyon}\\
Ecully, France \\
maxime.petit@ec-lyon.fr}
\and
\IEEEauthorblockN{2\textsuperscript{nd} Emmanuel Dellandrea}
\IEEEauthorblockA{\textit{LIRIS, CNRS UMR 5205} \\
\textit{Ecole Centrale de Lyon}\\
Ecully, France \\
emmanuel.dellandrea@ec-lyon.fr}
\and
\IEEEauthorblockN{3\textsuperscript{rd} Liming Chen}
\IEEEauthorblockA{\textit{LIRIS, CNRS UMR 5205} \\
\textit{Ecole Centrale de Lyon}\\
Ecully, France \\
liming.chen@ec-lyon.fr}
}

\maketitle
\thispagestyle{fancy}

\begin{abstract}
In robotics, methods and softwares usually require optimizations of hyperparameters in order to be efficient for specific tasks, for instance industrial bin-picking from homogeneous heaps of different objects. We present a developmental framework based on long-term memory and reasoning modules (Bayesian Optimisation, visual similarity and parameters bounds reduction) allowing a robot to use meta-learning mechanism increasing the efficiency of such continuous and constrained parameters optimizations. The new optimization, viewed as a learning for the robot, can take advantage of past experiences (stored in the \textit{episodic} and \textit{procedural} memories) to shrink the search space by using reduced parameters bounds computed from the best optimizations realized by the robot with similar tasks of the new one (\textit{e.g.} bin-picking from an homogenous heap of a similar object, based on visual similarity of objects stored in the \textit{semantic} memory). As example, we have confronted the system to the constrained optimizations of 9 continuous hyper-parameters for a professional software (Kamido) in industrial robotic arm bin-picking tasks, a step that is needed each time to handle correctly new object. We used a simulator to create bin-picking tasks for 8 different objects (7 in simulation and one with real setup, without and with meta-learning with experiences coming from other similar objects) achieving goods results despite a very small optimization budget, with a better performance reached when meta-learning is used (84.3\% vs 78.9\% of success overall, with a small budget of 30 iterations for each optimization) for every object tested (p-value=0.036).
\end{abstract}


%\vspace{-0.1cm}
\section{Introduction}

	\begin{figure}[!ht]
	%\vspace{-0.3cm}
        \centering
        \includegraphics[width=1.0\linewidth]{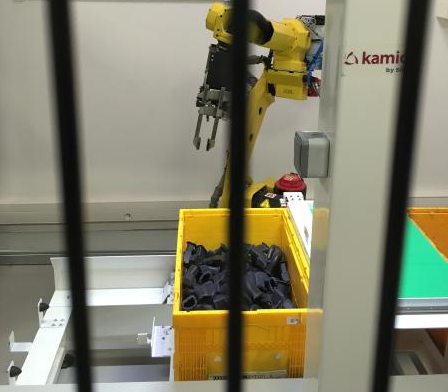}
    \caption{Real robotics setup with an industrial Fanuc robot for a grasping task from homogeneous highly cluttered heap of elbowed rubber tubes.}
    \label{fig-setup}
    %\vspace{-0.4cm}
    \end{figure}

\begin{figure*}[ht!]
%\vspace{-0.3cm}
\centerline{\includegraphics[width=0.9\linewidth]{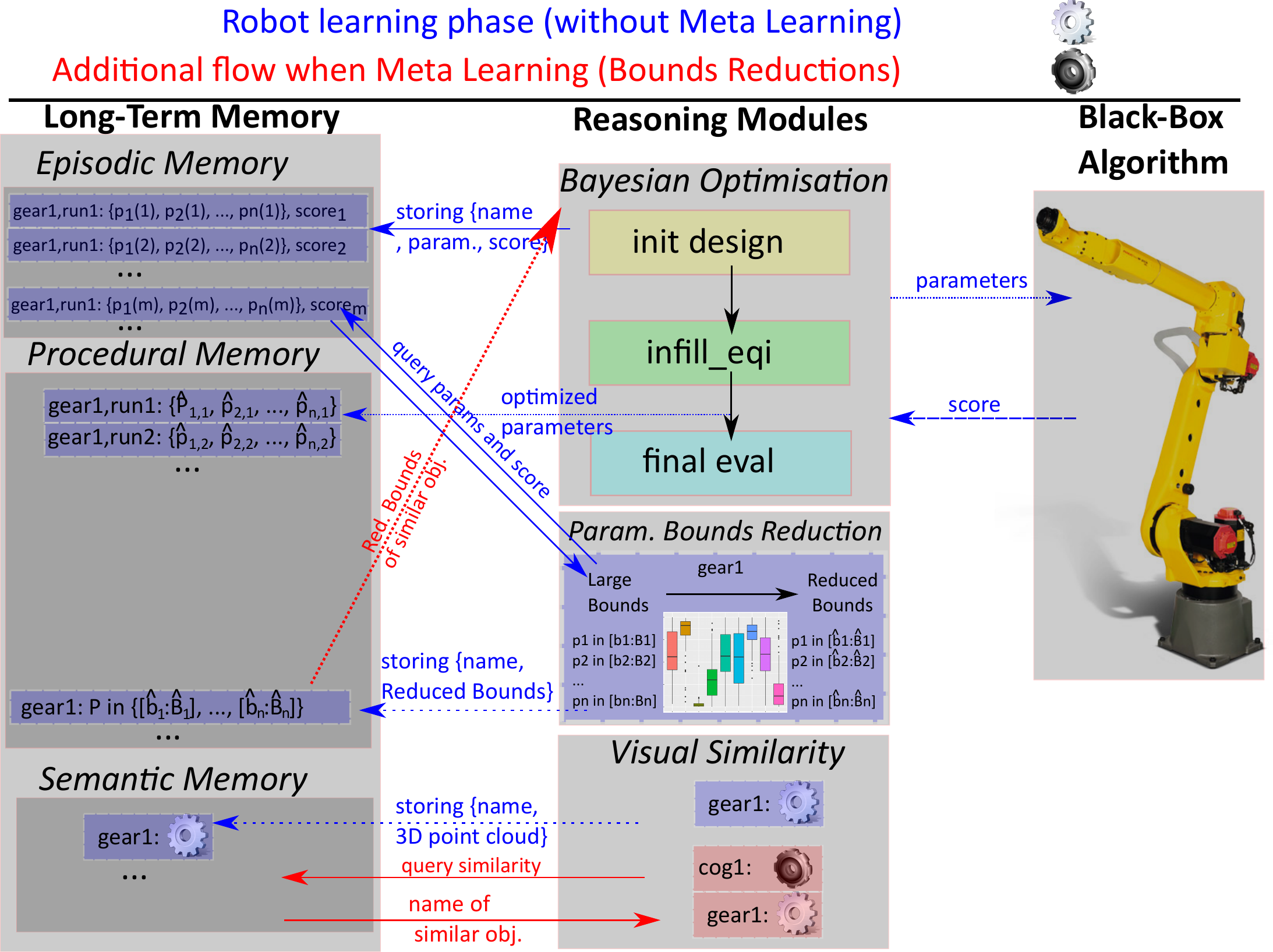}}
%\vspace{-0.3cm}
\caption{Architecture of the extended cognitive developmental framework, based on Long-Term Memory (with episodic, procedural and semantic memories) and Reasoning Modules (Bayesian Optimisation, Visual Similarity and the new Parameters Bounds Reduction) allowing a robot to learn how to grasp objects. This learning consists of guiding an efficient continuous hyper-parameters constrained optimization of black-box algorithm controlling the robot. The blue arrows represent the data flows during a learning phase without meta-learning (\textit{i.e.} without taking advantage of the Long-Term Memory, just storing the experiences). The red arrows shows the additional queries and exchanges of information during a learning phase with meta-learning, based on the visual similarity between objects the robot knows how to grasp and the new one.}
\label{fig-architecture}
%\vspace{-0.5cm}
\end{figure*}

In the field of robotics, many frameworks and algorithms require optional parameters settings in order to achieve strong performance (\textit{e.g.} Deep Neural Networks~\cite{snoek2012practical}, Reinforcement Learning~\cite{ruckstiess2010exploring}). Even if a human expert can manually optimized them, the task is tedious and error-prone, in addition to being costly in term of time and money when applied to the private industrial sector, in particular in situations where the hyper-parameters have to be defined frequently (\textit{e.g.} for each object to be manipulated or for each manipulation task). Optimization processes can be used to overcome these challenges on constrained numerical hyper-parameters search, such as Bayesian Optimization~\cite{mockus1989bayesian,mockus1994,brochu2010tutorial}. This method is especially suited where running the software (treated as black-box function) to be optimized will be expensive in time and will produce noisy score (the case for real robotics grasping applications). These methods are classically used before the deployment of the system \textit{in-situ}, or launched manually when needed: they are separated from the autonomous "life" of the robot's experience (\textit{i.e.} they are used offline). Therefore the optimizations are always starting from scratch (\textit{i.e.} \textit{cold-start}) because they are not taking advantage of the knowledge from previous experiences of the system (\textit{i.e.} \textit{warm-start}\cite{yogatama2014efficient}).

Our contribution consists of an enhanced version of the work from Petit \textit{et al.}~\cite{petit2018}: a developmental cognitive architecture providing a robot with a long-term memory and reasoning modules. It is allowing the robot to store optimization runs for bin-picking tasks using a professional grasping software, and utilize such experiences to increase the performance of new optimizations. In their initial works, when confronted to a new object for the bin-picking for which the grasping software parameters will have to be optimized, the robot is able to find a better solution faster with a transfer-learning strategy. This consists of extracting the best sets of parameters already optimized from a similar object and forcing the reasoning module to try it at the beginning of the optimization. Our contribution is the design of a meta-learning method for such optimization, in order to reduce the search space initially, thus avoiding unnecessary explorations in some areas. More specifically, we will use reduced parameters bounds that are extracted from the best previous optimization iterations of task or object that are similar to the new one, leading to a more efficient learning.

% Our contribution consist of a developmental cognitive architecture (composed of a long term memory and reasoning modules) allowing a robot to optimize by experience the parameters of a manipulation and/or vision algorithm (treated as black-box) where fine-tuning according to objects is needed. The learning procedure efficiency is increased by taking advantage of previous experiences (\textit{i.e.} past optimization of similar objects). The framework will be tested in both simulation and with real robot.

%\vspace{-0.1cm}
\section{Related Work}

Bayesian Optimization (BO) is a common method in the robotic field for optimizing quickly and efficiently constrained numerical parameters~\cite{lizotte2007automatic,calandra2016bayesian,yang2018learning}. In particular, Cully \textit{et al} implemented an extended version allowing a robot to quickly adjust its parametric gait after been damaged~\cite{cully2015robots} by taking advantages of previous simulated experiences with damaged legs. The best walking strategies among them were stored in a 6-dimensional behavioural grid (discretized with 5 values per dimension representing the portion of time of each leg in contact with the floor). We take inspiration from this work, where the behavioural space will be represented by the similarity between objects the robot will have to learn to manipulate.

The meta-learning concept of this work, focusing on reducing the initial search space of constrained numerical parameters optimization is inspired by the work of Maesani \textit{et al.}~\cite{maesani2014, maesani2015} known as the Viability Evolution principle. It consists, during evolutionary algorithms, of eliminating beforehand newly evolved agents that are not satisfying a viability criteria, defined as bounds on constraints that are made more stringent over the generations. This is forcing the generated agents to evolve within a smaller and promising region at each step, increasing the efficiency of the overall algorithm. We follow here the same principle by reducing the hyperparameters bounds based on past similar experience before the beginning of the optimization process, providing to it a smaller search space.

%\vspace{-0.1cm}
\section{Methodology}

The architecture of the cognitive robotics framework (see Fig.~\ref{fig-architecture}) is based upon the work of Petit et al.~\cite{petit2018}. It consists of the construction and exploitation with different reasoning capacities of a Long-Term Memory storing information in 3 sub-memories as described by Tulving~\cite{tulving1985memory}: 1) the \textit{episodic memory} storing data from personal experiences and events, then linked to specific place and time, 2) the \textit{procedural memory} containing motor skills and action strategies learnt during the lifetime and 3) the \textit{semantic memory} filled with facts and knowledge about the world. The developmental optimization with meta-learning will use this framework as followed: the Bayesian Optimization will provide all the data about its exploration and store them in the \textit{episodic memory}, with the optimized set of parameters stored in the \textit{procedural memory}. Parameters Bounds Reduction module will analyze the data for each task from the \textit{episodic memory} in order to compute reduced parameters bounds still containing the best values for each parameters. A Visual Similarity module will be able to compare the similarity between different tasks (\textit{e.g.} grasping an object $O_1$ and an object $O_2$) in order to have access to previous knowledge stored in the \textit{procedural memory} and linked to a known similar task when confronted to a new one. This will allow the robot to use a smaller search optimization space when trying to learn how to achieve a task A by using the reduced parameters bounds computed from a similar and already explored and optimized task B. %\textcolor{red}{The Bayesian Optimisation and Visual Similarity module are the same than the one detailed in our previous work (see~\cite{petit2018} for more details if needed) and therefore will only be shortly described here.}

%\vspace{-0.1cm}
\subsection{Bayesian Optimisation module}\label{BO}

We have chosen Bayesian Optimization as method for constrained optimization process of the robotic algorithm black-box, implemented using the R package \textit{mlrMBO}~\cite{mlrMBO} with Gaussian Process as surrogate model. A BO run optimizes a number of parameters with iterations (\textit{i.e.} trials) where the set of parameters is selected (and tested) differently depending on the current phase, out of 3, of the process:
\begin{itemize}
	\item \textit{"initial design"}: selecting points independently to draw a first estimation of the objective function. 
    \item Bayesian search mechanism (\textit{"infill eqi"}), balancing exploitation and exploration. It is done by extracting the next point from the acquisition function (constructed from the posterior distribution over the objective function) with a specific criteria. We have chosen to use the Expected Quantile Improvement (EQI) criteria from Pichney \textit{et al.}\cite{Picheny2013} because the function to optimize is heterogeneously noisy. EQI is an extension of the Expected Improvement (EI) criteria where the improvement is measured in the model rather than on the noisy data, and so is actually designed to deal with such difficult functions. 
    \item final evaluation (\textit{"final eval"}), where the best predicted set of hyper-parameters (prediction of the surrogate, which reflects the mean and is less affected by the noise) is used several times in order to provide an adequate performance estimation of the optimization.
\end{itemize}

%\vspace{-0.1cm}
\subsection{Memory}

Similarly to others implementations of a long-term memory system~\cite{pointeau2014,petit2016}, the experience and knowledge of the robot are stored in a PostgreSQL database. The \textit{episodic} memory stores each experience of the robot, and consists for this work of the information available after each iteration $i$ of the Bayesian Optimization's run $r$: the label of the task (\textit{e.g.} the name of the object for which the robot has to optimize parameters in order to manipulate it), the set of $m$ hyper-parameters tested $\{p_1(i), p_2(i), ..., p_m(i)\}$ and the corresponding score $s_i$ obtained with such setup. The \textit{semantic memory} is filled and accessed by the Visual Similarity module and contains the visual information about the objects that the robot used during its optimization runs, and are stored as point clouds. The \textit{procedural memory} is composed by 2 types of data: 1) optimized sets of parameters of each run of each object are stored by the Bayesian Optimisation module, in order to be quickly loaded by the robot if needed, and 2) reduced parameters bounds for each object, corresponding of constrained boundaries for each parameters values obtained when looking at the parameters values distribution from the best iterations of a specific task/object. This information is pushed here by the Parameters Bounds Reduction module, that we will describe later.

%\vspace{-0.1cm}
\subsection{Visual Similarity module}\label{VS}

The Visual Similarity module is retrieving the most similar object from the \textit{semantic} memory (\textit{i.e.} CAD model of known object, meaning the robot has already optimized the corresponding parameters) where confronted to CAD models of a new objects to be optimized. It is based on an extension of the deep learning method for 3D classification and segmentation PointNet~\cite{pointnet} which provides a numerical metrics for the similarity between 2 objects as the distance of the 1024 dimensions global features from the models. The most similar object corresponds to the minimal distance.

%\vspace{-0.1cm}
\subsection{Meta Learning: Parameters Bounds Reductions}

	\begin{figure}[!ht]
	%\vspace{-0.3cm}
        \centering
        \includegraphics[width=1.0\linewidth]{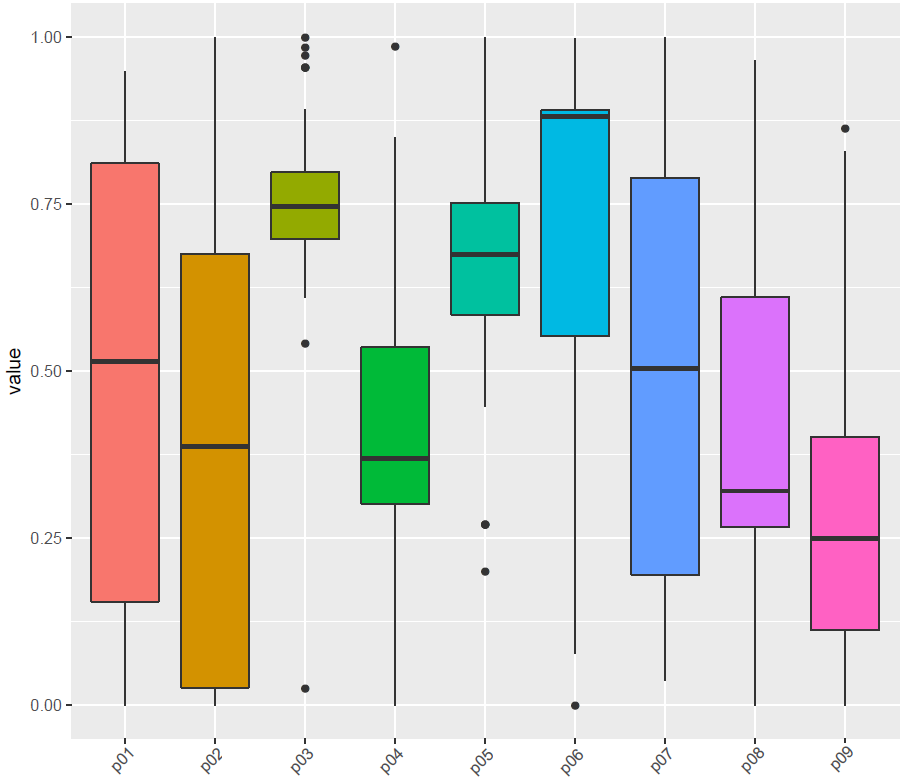}
    \caption{Distribution of the scaled values of 9 parameters from the best 35\% optimizations iterations of the object to be grasped called \textit{m782}. Some parameters have a uniform [0:1] distribution (\textit{e.g.} p1) but some do not and their median is either around 0.5 (\textit{e.g.} p7), higher (\textit{e.g.} p5) or smaller (\textit{e.g.} p9). See Table~\ref{tab-bounds} for the corresponding new reduced parameter bounds.}
    \label{fig-param-distrib}
    %\vspace{-0.4cm}
    \end{figure}
    
    \begin{algorithm}
     \caption{Algorithm for bounds Reduction}
         \begin{algorithmic}[1]
             \renewcommand{\algorithmicrequire}{\textbf{Input:}}
             \renewcommand{\algorithmicensure}{\textbf{Output:}}
             \REQUIRE All iterations of all runs for object $O$ with scaled parameters values ($\in [0:1]$)
             \ENSURE  New reduced bounds for object $O$
             %\\ \textit{Initialisation} :
              \STATE Select $I_{n}(O)$ the n\% best iterations for $O$
             %\\ \textit{LOOP Process}
              \FOR {each parameters $p_j(O)$}
              \STATE Compute $p_{dm}$, p-value of Dudewicz-van der Meulen test for uniformity for $p_j(O)$ values from $I_{n}(O)$
              \IF {($p_{dm} < \alpha_{dm}$)}
              \STATE Compute $p_w$, p-value of Wilcoxon test (H0: $\mu=0.5$)
                \IF {($p_{w} < \alpha_w$ and median($p_j(O)$)$> 0.5$)}
                \STATE Increase lower bound for $p_j(O)$ to the 5\% percentile of $p_i(O)$ values from $I_{n}(O)$
                \ELSIF {($p_{w} < \alpha_w$ median($p_j(O)$)$< 0.5$)}
                \STATE Reduce upper bound for $p_j(O)$ to the 95\% percentile of $p_j(O)$ values from $I_{n}(O)$
                \ELSE 
                \STATE Reduce upper \& increase lower bounds for $p_i(O)$
                \ENDIF
              \ENDIF
              \ENDFOR
             \RETURN Modified Parameters bounds
         \end{algorithmic} 
         \label{alg-bounds}
     \end{algorithm}

    \begin{figure*}[ht!]
    %\vspace{-0.3cm}
    \centerline{\includegraphics[width=1.0\linewidth]{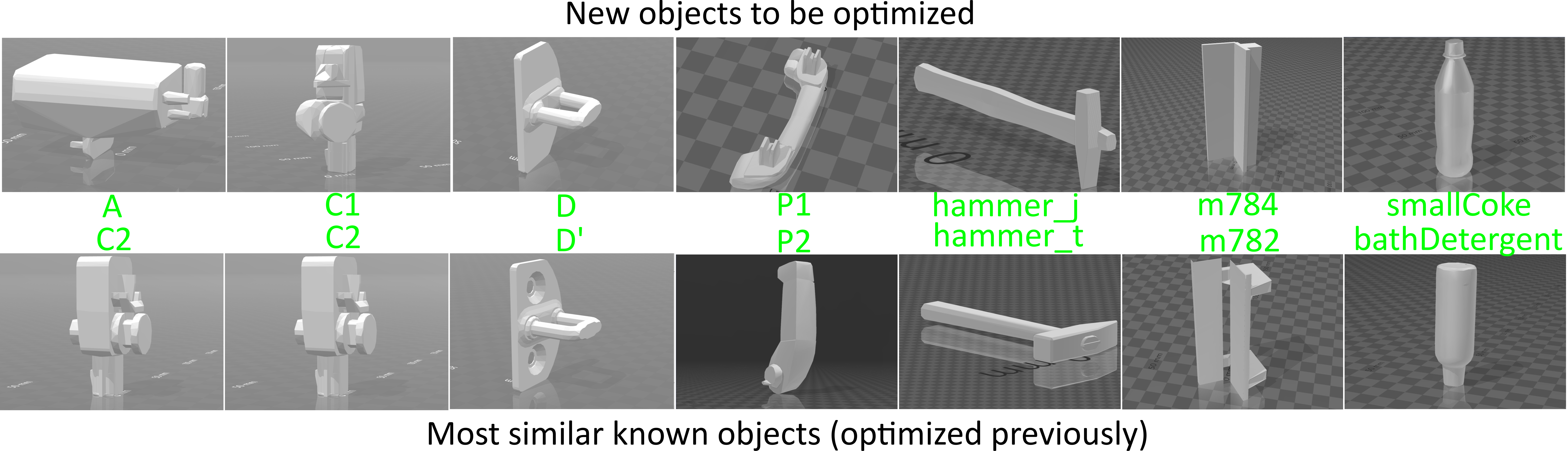}}
    %\vspace{-0.3cm}
    \caption{CAD models of the simulated objects}
    \label{fig-obj}
    %\vspace{-0.5cm}
    \end{figure*}

The Meta Learning aspect is realized with the use of reduced, more adequate, promising and efficient parameters bounds when launching the constrained optimization of a novel task (\textit{i.e.} bin-picking a new object), using the reduced parameters bounds extracted from the experience of the robot with bin-picking a similar object to the new one. When looking at the distribution of the parameters values explored during the iterations that provided the best results, an efficient parameters bounds would provide roughly a uniform distribution of the parameters values among the best iteration, meaning that they are many parameters values within that provide good results. On the opposite, a very thin distribution means that a huge part of the search landscape for the parameters are sub-optimized and will cost optimization budget to be explored futilely. We want then to reduce the parameters bounds in order to force the optimization process to focus on the more promising search space. We describe here how the module is able to reduced the parameters bounds from past optimization of an object O, summarized in Alg.~\ref{alg-bounds} in order to increase the efficiency of future optimization runs for the same or similar object.

First, the module is checking the \textit{episodic} memory of the robot to retrieve every results of past optimization iterations for the object O, $I(O)$. Among them, we only keep the iterations that provided the best results, filtering to have the n\% best remaining and obtain $I_{n}(O)$, a subset of $I(O)$. Then the module will analyze the distribution of every parameters $p_j$ explored for the object O and scaled in [0:1], where an example of such distribution is shown in Fig.~\ref{fig-param-distrib} under the form of boxplots. For each parameter, we check the uniformity of the distribution in [0:1] using the Dudewicz-van der Meulen test~\cite{dudewicz1981}, an entropy-based test for uniformity over this specific distribution. If the p-value $p_{dm}$ is below the alpha risk $\alpha_{dm}$, we can reject the uniformity hypothesis for the current distribution: we can eliminate some range values for the parameter. However, it can goes several ways: we can lower the upper bounds, increasing the lower bounds, or doing both. This decision will be based on the result on a non-parametric (we cannot assume the normality of the distribution) one-sample Wilcoxon signed rank test against an expected median of $\mu = 0.5$ producing a p-value $p_w$ and using another alpha risk $\alpha_{w}$. If the $p_w < \alpha_{w}$ we can reject the hypothesis that the distribution is balanced and centered around 0.5. If that is the case and the distribution is not uniform, that means that both bounds can to be reduced (lowering the upper bounds and increasing the lower one). If not, that means the distribution is favoring one side (depending on the median value) and only the bounds from the opposite side will be more constrained: the lower bounds will be increased if the median is greater than 0.5, or the upper bounds will be smaller if the median is lower than 0.5. The bounds are modified to the $x^{th}$ percentile value of the parameters for the lower bounds and to the $X^{th}$ percentile for the upper bounds, with $0 \leq x < X \leq 1$. Eventually, they are stored in the \textit{procedural memory} and linked to their corresponding object, in order to be easily accessible in the future and used by future optimization process instead of the default and larger parameters bounds.

     % test is an entropy-based test for uniformity over a [0:1] distribution 
    %of $x_1, x_2, ..., x_n$ values, based on the following statistic:
    % \begin{equation}
    %     H(m,n) = -\frac{1}{n}\sum_{i=1}^{n}\log_{2}\frac{n}{2m}(x_{(i+m)}-x_{(i-m)})
    % \end{equation}
    % where $m \leq \frac{n}{2}$

    %\textcolor{red}{Parameters are scaled in [0:1] before and after the alg.}

% 	\begin{figure}[!htb]
%     \begin{minipage}{.65\columnwidth}
%         \centering
%         \includegraphics[width=0.95\linewidth]{./img/boxplot_CapteurDetLat_75.pdf}
%     \end{minipage}%
%     \begin{minipage}{0.325\columnwidth}
%         \centering
%         \begin{center}
%         \begin{scriptsize}
%             \begin{tabularx}{\columnwidth}{|C|X|X|}
%             \hline
%             Par. & Default & C2$\_$25 \\ \hline \hline
%             p01 & [-20:20] & [-20:20] \\ \hline
%             p02 & [5:15]   & [\textbf{8.3}:15] \\ \hline
%             p03 & [16:100] & [\textbf{46}:\textbf{92}] \\ \hline
%             p04 & [5:30]   & [5:30] \\ \hline
%             p05 & [5:30]   & [\textbf{13}:30] \\ \hline
%             p06 & [5:40]   & [\textbf{24}:\textbf{37}] \\ \hline
%             p07 & [30:300] & [\textbf{100}:\textbf{220}] \\ \hline
%             p08 & [5:20]   & [5:\textbf{15}] \\ \hline
%             p09 & [1:10]   & [\textbf{2.5}:\textbf{9}] \\ \hline
%             \end{tabularx}
%         \end{scriptsize}
%         \end{center}
%     \end{minipage}
%     \caption{test}
%     \end{figure}

%\vspace{-0.2cm}

\section{Experiments}

    % \begin{figure}[!htb]
    % %\vspace{-0.3cm}
    % \centering
    %     \includegraphics[width=0.9\linewidth]{./img/ICRA2020_newobj.pdf}
    % \caption{CAD models of the new objects (see \cite{petit2018} for the others)}
    % \label{fig-new-obj}
    % \vspace{-0.4cm}
    % \end{figure}

The experiment setup is similar to the describe in ~\cite{petit2018} allowing to compare some of their results with ours. We are indeed aiming at optimizing some parameters of a professional software called Kamido\footnote{http://www.sileane.com/en/solution/gamme-kamido} (from Sileane) that we are treating as a black-box. The parameters are used by Kamido to analyze RGB-D images from a fixed camera on top of a bin and extract an appropriate grasping target for an industrial robotic arm with parallel-jaws gripper in a bin-picking task from an homogeneous heap (\textit{i.e.} clutter composed by several instances of the same object).

We use real-time physics PyBullet simulations where objects are instantiated from Wavefront OBJ format on which we apply a volumetric hierarchical approximate convex decomposition~\cite{vhacd16}. The function to be optimized will be the percentage of success at bin-picking, where an iteration of the task consist of 15 attempts to grasp cluttered objects in the bin and to release the catch in a box. We also introduce a partial reward (0.5 instead of 1) when the robot is grasping an object but fails to drop it into the deposit box.

To be able to compare each learning run with the same learning condition, we authorize a finite budget for the BO process of 35 iterations, decomposed as follows: 10 for the \textit{"initial design"}, 20 for the core BO process and 5 as repetitions of the optimized set of parameters in order to provide a more precise estimation of the performance. As opposed to the experiment done in ~\cite{petit2018}, we decided to constrain more the learning setup, providing only 30 (10+20) iterations instead of 68 (18+50). Indeed, the learning curve seemed to flattened around this number of iterations in their work, so we wanted to compare the quality of the optimization at an earlier stage. For the bounds reduction algorithm, we use a selection of the best 35\% iterations for each object thus allowing a good range of potential efficient set of parameters from a very noisy objective function, and alpha risk of 0.15 for both the Dudewicz-van der Meulen and Wilcoxon tests (\textit{i.e.} $\alpha_{dm} = \alpha_w = 0.15$). The percentile used for the bounds reductions are x=0.05 and X=0.95 in order to discard any potential outliers that might otherwise forbid a strong reduction in boundaries.

The other aspect of the setup are unchanged. Indeed, during the initial design phase, the set of parameters are selected using a Maximin Latin Hypercube function~\cite{lhsMaximin} allowing a better exploration  by maximizing the minimum distance between them. The kernel for the GP is the classic Matern 3/2 and the criteria for the bayesian search mechanism on the acquisition function is an EQI with a quantile level of $\beta = 0.65$. The infill criterion is optimized using a stochastic derivative-free numerical optimization algorithm known as the Covariance Matrix Adapting Evolutionary Strategy (CMA-ES) ~\cite{cmaes1,cmaes2} from the package \textit{cmaes}.

For the experiments presented in this work, we used some objects from ~\cite{petit2018}, namely the reference A, C1, C2, D and D' in order to compare the performance of the method with a smaller learning budget as explained earlier. We also introduce new objects, some from a CAD database of real industrial reference (P1 and P2), and some from other common databases, such as \textit{hammer$\_$t} and \textit{hammer$\_$j} from turbosquid, \textit{m782} and \textit{m784} from Princeton Shape Benchmark\cite{shilane2004}, and \textit{bathDetergent} and \textit{cokeSmallGrasp} from KIT\cite{kasper2012}. New objects are shown in Fig.~\ref{fig-obj}, along the objects (C2, C2, D', P2, \textit{hammer$\_$t} and \textit{m782}) that has been optimized previously by the robot and that is the most similar, using the Visual Similarity module.
%m548?

The experiments will consist of the optimization process for 7 objects (A, C1, D, P1, \textit{hammer$\_$j}, \textit{m784} and \textit{cokeSmall}) taken from 4 different object databases) when the method has been applied 6 times independently (\textit{i.e.} runs) with 2 conditions: one optimization without any prior knowledge use, and one using meta-learning. This last condition involves retrieving the most similar and already optimized object known by the robot when confronted to the optimization of a new unknown object. Then the robot extracts the reduced boundaries of the best set of parameters it already tried with the similar object (the best 35\% set of parameters) using the appropriate reasoning module described earlier. It then constrains the parameters values with these new reduced bounds during the optimization process. The reduced parameters bounds of each object similar to the references are presented in Table~\ref{tab-bounds}.

   \begin{table}
        \caption{Bounds for each parameter to be optimized, with the larger "Default" and the reduced bounds obtained from several objects using the Parameters Bounds Reductions module.}
        \vspace{-0.1cm}
        \begin{scriptsize}
            \begin{tabularx}{\columnwidth}{|@{\hskip2pt}L@{\hskip2pt}|@{\hskip2pt}X@{\hskip2pt}|@{\hskip2pt}X@{\hskip2pt}|@{\hskip2pt}X@{\hskip2pt}|@{\hskip2pt}X@{\hskip2pt}|@{\hskip2pt}X@{\hskip2pt}|@{\hskip2pt}X@{\hskip2pt}|@{\hskip2pt}X@{\hskip2pt}|@{\hskip2pt}X@{\hskip2pt}|@{\hskip2pt}X@{\hskip2pt}|}
            \hline
            Obj. & p1 & p2 & p3 & p4 & p5 & p6 & p7 & p8 & p9 \\ \hline \hline
            Def. & -20:20 & 5:15 & 16:100 & 5:30 & 5:30 & 5:40 & 30:300 & 5:20 & 1:10 \\ \hline
            C2 & -20:20 & \textcolor{red}{\textbf{8}}:15 & \textcolor{red}{\textbf{46}}:\textcolor{red}{\textbf{92}} & 5:30 & \textcolor{red}{\textbf{13}}:30 & \textcolor{red}{\textbf{24}}:\textcolor{red}{\textbf{37}} & \textcolor{red}{\textbf{100}}:\textcolor{red}{\textbf{220}} & 5:\textcolor{red}{\textbf{15}} & \textcolor{red}{\textbf{3}}:\textcolor{red}{\textbf{9}} \\ \hline
            D' & \textcolor{red}{\textbf{-18}}:\textcolor{red}{\textbf{10}} & 5:15 & \textcolor{red}{\textbf{49}}:\textcolor{red}{\textbf{99}} & \textcolor{red}{\textbf{8}}:\textcolor{red}{\textbf{23}} & 5:30 & 5:40 & 30:300 & \textcolor{red}{\textbf{8}}:20 & \textcolor{red}{\textbf{2}}:\textcolor{red}{\textbf{8}}\\ \hline
            P$\_$2 & -20:20 & \textcolor{red}{\textbf{6}}:\textcolor{red}{\textbf{14}} & \textcolor{red}{\textbf{20}}:\textcolor{red}{\textbf{69}} & 5:30 & 5:30 & \textcolor{red}{\textbf{18}}:\textcolor{red}{\textbf{37}} & \textcolor{red}{\textbf{114}}:\textcolor{red}{\textbf{267}} & 5:\textcolor{red}{\textbf{19}} & 1:10 \\ \hline
            %Can. & \textbf{-18}:\textbf{18} & \textbf{6}:\textbf{14} & \textbf{18}:\textbf{64} & 5:30 & \textbf{13}:\textbf{29} & 5:40 & 30:300 & 5:20 & 1:10 \\ \hline
            ham$\_$t & -20:20 & 5:15 & \textcolor{red}{\textbf{46}}:100 & \textcolor{red}{\textbf{8}}:30 & 5:30 & \textcolor{red}{\textbf{17}}:40 & 30:300 & 5:20 & 1:10 \\ \hline
            m782 & -20:20 & 5:15 & \textcolor{red}{\textbf{68}}:\textcolor{red}{\textbf{96}} & \textcolor{red}{\textbf{7}}:\textcolor{red}{\textbf{23}} & \textcolor{red}{\textbf{12}}:30 & \textcolor{red}{\textbf{9}}:\textcolor{red}{\textbf{37}} & 30:300 & 5:\textcolor{red}{\textbf{19}} & 1:\textcolor{red}{\textbf{8}} \\ \hline
            bathDet. & \textcolor{red}{\textbf{-15}}:\textcolor{red}{\textbf{9}} & \textcolor{red}{\textbf{9}}:15 & \textcolor{red}{\textbf{69}}:100 & \textcolor{red}{\textbf{10}}:30 & \textcolor{red}{\textbf{18}}:30 & \textcolor{red}{\textbf{27}}:40 & 30:\textcolor{red}{\textbf{276}} & 5:20 & 1:10 \\ \hline
            \end{tabularx}
        \end{scriptsize}
    \label{tab-bounds}
    %\vspace{-0.3cm}
    \end{table}

%\vspace{-0.1cm}
\section{Results}

\begin{figure}[!htb]
        %\vspace{-0.3cm}
        \centering
        \includegraphics[width=1.0\linewidth]{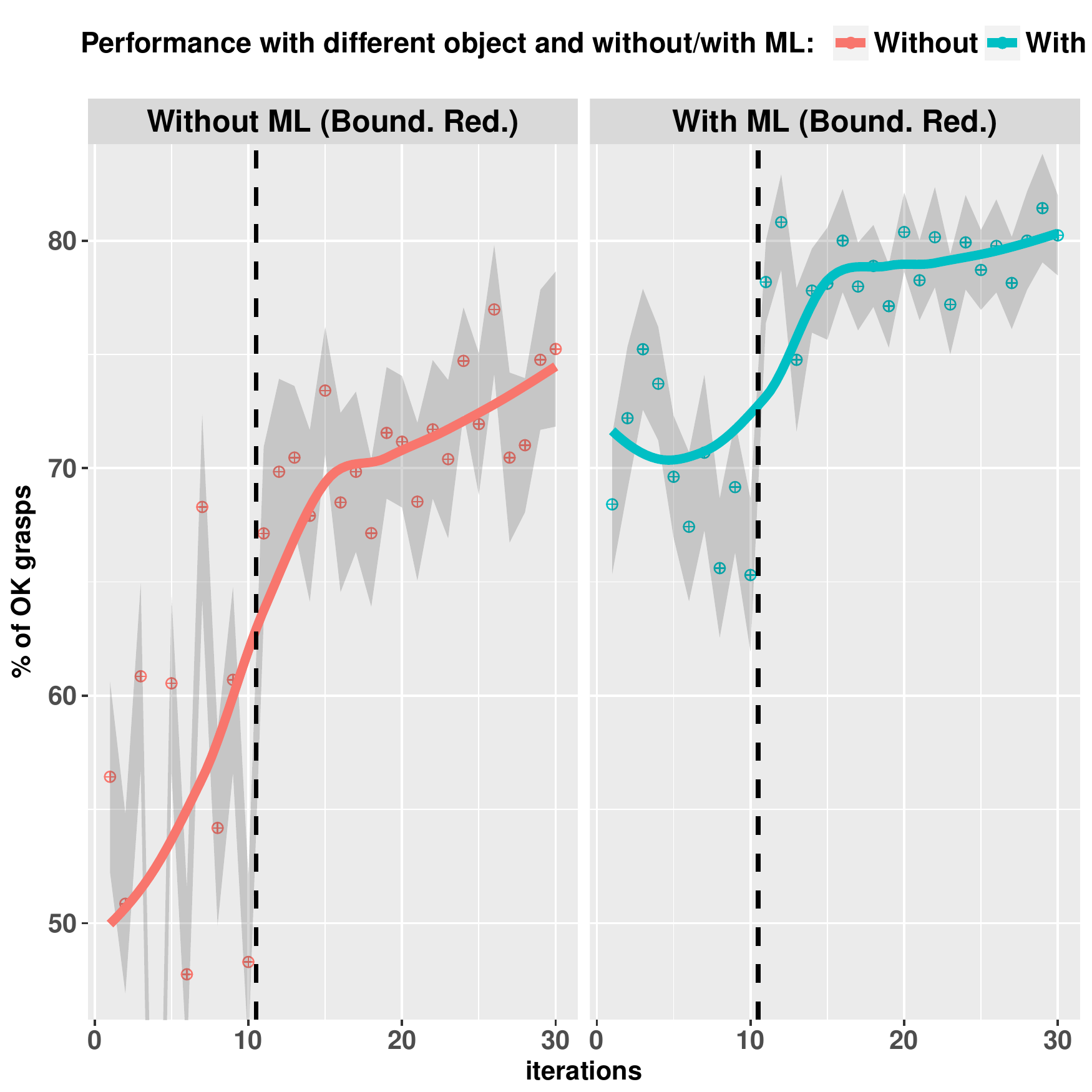}
    \caption{Performance for each iteration (all objects) of the optimization runs, during the \textit{initial design} (Iteration 1:10, Left of the vertical dotted line) and \textit{infill eqi criteria} phase (Iteration 11-30, Right of the dotted line). Crossed circles are means among all runs at each iteration, while the grey area is the standard deviation. Curves corresponds to a smoothing among the points, using the non-parametric LOcally weighted redgrESSion (\textit{i.e.} loess) method.}
    \label{fig-init-eqi-curve}
    %\vspace{-0.4cm}
    \end{figure}

In this section, we present the results from the experiments, focusing first on the performance during the optimization process, at both \textit{initial design} and \textit{infill eqi criteria} phase, with the Fig.~\ref{fig-init-eqi-curve}. We can see that using the meta-learning (\textit{i.e.} using prior information about the performance of set of parameters from similar object to the new one) allows the optimization process to have a \textit{warmstart} during the \textit{initial design} phase with a mean performance of already more than 75\% compared to $\sim$65\% when the parameters bounds are not restricted. It means that the algorithm process is avoiding spending optimization budget to explore parameters values that are inside the default bounds, but outside the bounds of interests from similar object, thus exploring un-optimized parameters values. This leads to a search space with more promising areas densities that the Bayesian Optimization process is able to explore more efficiently during the \textit{infill eqi criteria} phase.

\begin{figure}[!htb]
	%\vspace{-0.3cm}
        \centering
        \includegraphics[width=1.0\linewidth]{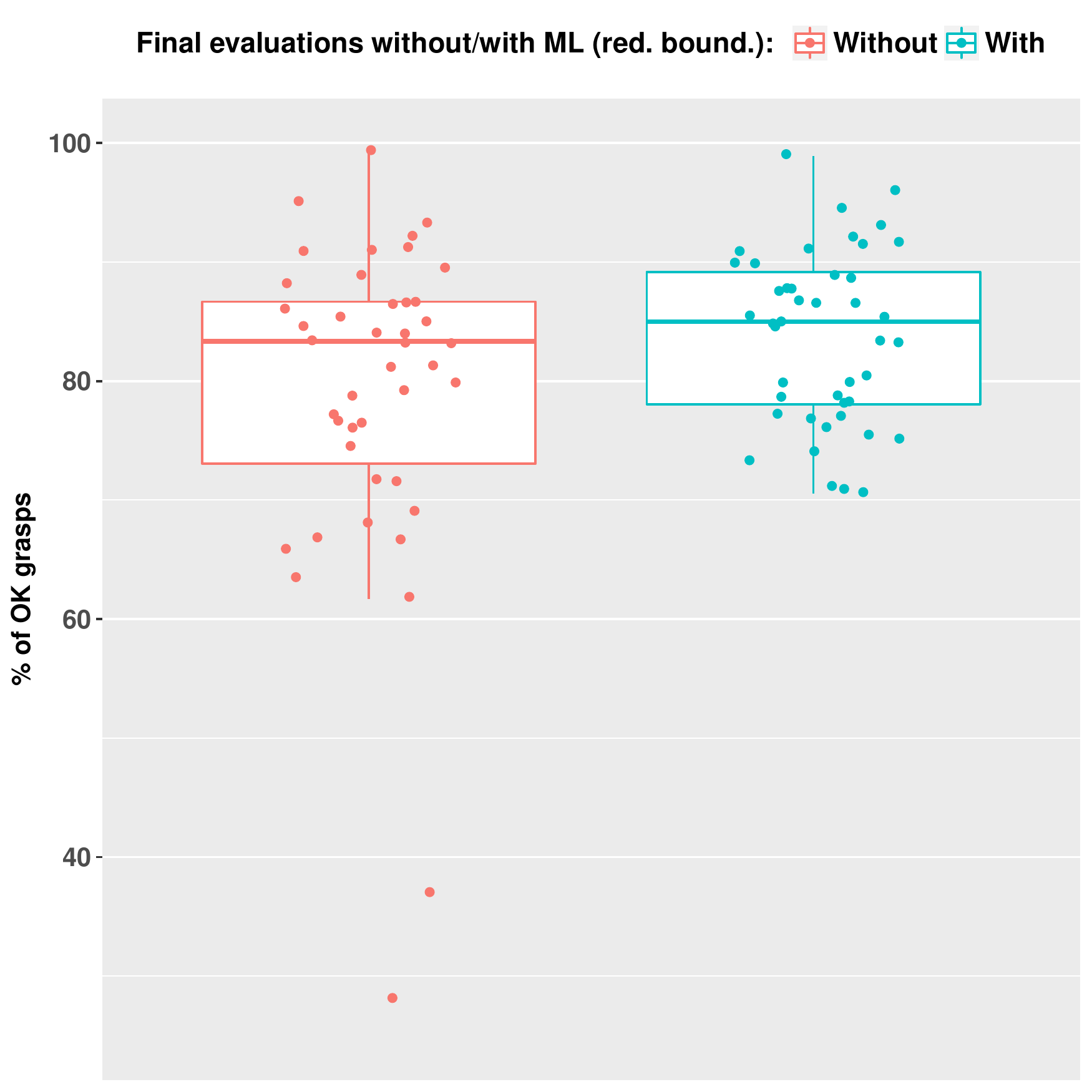}
    \caption{Boxplot of the final performance after Bayesian Optimization on all objects for all runs, without and with meta-learning (Parameters Bounds Reduction applied to new objects from the bounds of a similar optimized object). Each dot is the mean final performance after an optimization run.}
    \label{fig-final-box}
    %\vspace{-0.3cm}
    \end{figure}

We then look at the final performances of every runs for every objects, split in two sets (without and with meta-learning) shown in Fig.~\ref{fig-final-box}. The mean performance overall increases from 78.9\% (Q1: 73.1, median: 83.3, Q3:86.7) without the bounds reduction step to 84.3\% (Q1: 78.1, median: 85, Q3:89.2) when the Bayesian Optimization is using meta-learning (Wilcoxon test). In addition, the worst performance after optimization among every runs and objects, even with a very short learning budget (30 iterations to optimize 9 continuous hyper-parameters), is at a decent 70.6\% when using this meta-learning technique (vs 28.3\% otherwise).    %Wilcoxon test, p-value$=7.4.10^{-4}$, mean performance without 74.17\% vs 82.36\% with Meta-Lerning (all final eval iterations of all runs) %(0.056 p-value wilcox if mean perfo for each runs)

Detailed and numerical results of the experiments, split among all objects, are shown in Table~\ref{tab-res}. First, we can compare the performance of the optimization method for object A, C1 and D at an earlier stage (after 30 learning iteration instead of 68) than the experiments from ~\cite{petit2018}. We indeed achieved similar performance for these objects under this harsher experiment design but with meta-learning, with respectively a mean success among all runs of 75.9\%, 79.4\% and 89.4\% (30 iterations learning) vs 76.1\%, 81.3\% and 87.3\% (68 iterations learning).

\begin{figure}[!htb]
        %\vspace{-0.3cm}
        \centering
        \includegraphics[width=1.0\linewidth]{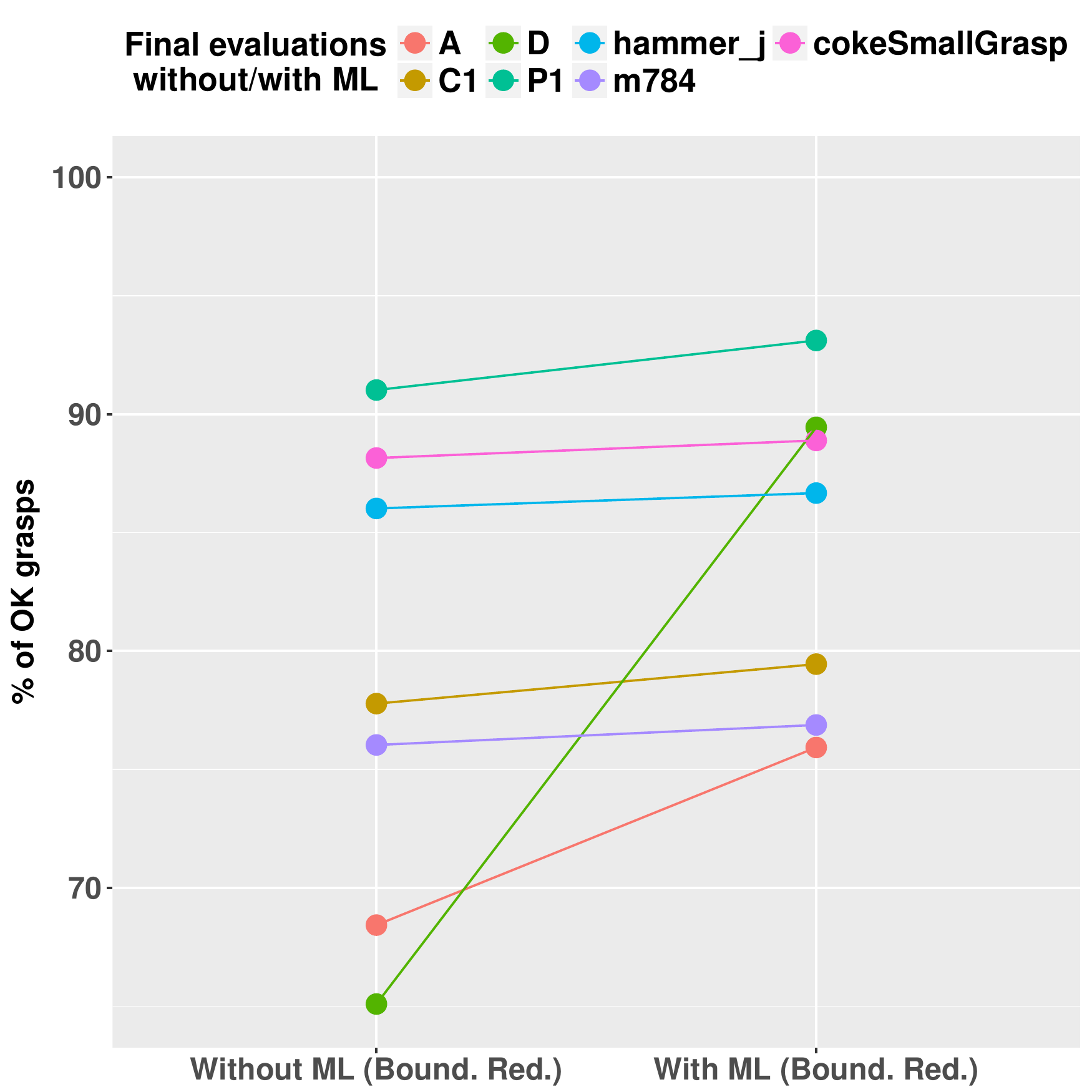}
    \caption{Final mean performance of all runs, grouped by objects and paired on both conditions: without meta-learning and with meta-learning. This shows the systematic gain of performance when using meta-learning strategy, with a greater benefit where the initial performance was lower (object D and A)}
    \label{fig-paired-mean}
    %\vspace{-0.3cm}
    \end{figure}

Looking at every object's performance, shown also in a paired graph from Fig.~\ref{fig-paired-mean} , We can also clearly see the benefit of using the meta-learning method during the optimization process, with a better mean performance for every object among all the runs, leading to a significantly better score (paired sampled Wilcoxon test p-value=0.031). Table~\ref{tab-res} also shows that worst performance is anyhow always better (at least $>70.6\%$) when using the meta-learning, providing a higher minimum expected performance (paired sampled Wilcoxon test p-value=0.031). Overall, it seems that the robot is benefiting more from the meta-learning when the task is more difficult (\textit{i.e.} when percentage of success is overall lower) like with objects A and D, with a lower success score with BO only of respectively 68.4\% and 65.1\%) and the constrained search space allows the Bayesian Optimization to be more efficient and find promising parameters sooner, and for each run. However, the Bayesian Optimisation can still be efficient even without meta-learning as seen from the performance of the best runs, however the optimization are less reliable: most runs will not be as efficient as with meta-learning.

\begin{table}[h]
\vspace{-0.1cm}
\begin{center}
\caption{Optimization Results with/without Meta Learning - Comparison with \cite{petit2018} using Budget of 68 iterations vs 30 here and Transfer Learning instead of Meta Learning \vspace{-0.1cm}}
\begin{tabularx}{\columnwidth}{|@{\hskip2pt}L|@{\hskip2pt}c@{\hskip2pt}|X|@{\hskip2pt}c@{\hskip2pt}|@{\hskip2pt}c@{\hskip2pt}|}
\hline
Reference & Budget &$\%$ succes all run & $\%$ succes &$\%$ succes \\
& & mean$\pm$sd, median & (worst run) & (best run) \\\hline \hline
A \cite{petit2018} & 68 & 65.47$\pm$27.3, 73.3 & - & 78.9 \\ \hline
A & 30 &68.4$\pm$7.09, 66.4 & 61.7 & 81.1 \\ \hline
A\_ML\_C2  & 30 &75.9$\pm$2.37, 75.8 & 73.3 & 80.0 \\ \hline
A\_TL\_C2 \cite{petit2018} & 68 & 76.1$\pm$10.19, 76.7 & - & 82.8 \\ \hline \hline
C1 \cite{petit2018}  & 68 &78.95$\pm$10.87, 80 & - & 83.9  \\ \hline
C1  & 30 &77.6$\pm$6.00, 77.5 & 68.3 & 85.0  \\ \hline
C1\_ML\_C2  & 30 & 79.4$\pm$5.44, 79.4 & 70.6 & 85.0  \\ \hline 
C1\_TL\_C2 \cite{petit2018} & 68 &81.3$\pm$11.04, 80 & - & 82.5  \\ \hline \hline
D \cite{petit2018}  & 68 &86.9$\pm$9.45, 86.67 & - & 91.1  \\ \hline
D  & 30 &65.1$\pm$25.7, 76.4 & 28.3 & 88.3  \\ \hline
D\_ML\_D'  & 30 &  89.4$\pm$6.78, 90 & 78.9 & 96.1  \\ \hline
D\_TL\_D' \cite{petit2018} & 68 &87.3$\pm$7.44, 86.7 & - & 90.6  \\ \hline \hline \hline
P1  & 30  & 91.0$\pm$6.06, 91.4 & 83.3 & 99.4   \\ \hline
P1\_ML\_P2 & 30 & 93.1$\pm$3.25, 91.7 & 91.1 & 98.9 \\ \hline \hline
ham$\_$j  & 30 & 86.0$\pm$4.8, 84.7 & 80.0 & 92.2  \\ \hline
ham$\_$j\_ML\_ham$\_$t & 30 & 86.7$\pm$2.06, 86.7 & 83.3 & 90.0 \\ \hline \hline
m784  & 30 &76.0$\pm$6.65, 76.7 & 66.7 & 86.7   \\ \hline
m784\_ML\_m782 & 30  & 76.9$\pm$4.27, 77.8 & 71.1 & 83.3  \\ \hline \hline 
coke  & 30 &88.1$\pm$2.69, 87.8 & 84.4 & 91.1   \\ \hline
coke\_ML\_detergent & 30  & 88.9$\pm$3.06, 88.9 & 85.6 & 93.3  \\ \hline
% Overall & 74.17 & 59.3& \\ \hline
% Overall\_ML & 82.36 & &\\ \hline
\end{tabularx}
\label{tab-res}
\end{center}
%\vspace{-0.5cm}
\end{table}

We have also implemented our architecture on a real robotic arm Fanuc, however the specific version of the robot (M20iA/12L vs M10iA12), the end-effector parallel-jaws gripper and the environmental setup (See Fig.~\ref{fig-setup}) is different than the one used in ~\cite{petit2018}, so direct comparison is not possible. In addition, because we used non-deformable object in simulation, we wanted to try with a real soft-body object in order to check if the method can obtain good results with such physical property. Therefore, we created an homogenous heap of highly cluttered elbowed rubber tube pieces as a test. With the 30 iterations budget runs, we have observed again a benefit of the meta-learning feature, with an increase from 75.6\% of mean performance with the real robot (sd=5.46, min=70.6, max=82.8) without meta-learning, to 84.6\% (sd=2.5, min=82.2, max=87.2) with meta-learning.

\section{Conclusion and Future Work}

This work explored how a robot can take advantage of its experience and long-term memory in order to utilize a meta-learning method and enhance the results of Bayesian Optimization algorithm for tuning constrained and continuous hyper-parameters, in bin-picking objects type of tasks (6 different objects extracted from 3 different shape objects database). With a very small fixed optimization budget of 30 trials, we are able to optimize 9 continuous parameters of an industrial grasping algorithm and achieve good performance, even with a very noisy evaluation function as encountered during this task. The meta-learning method, based on the reduction of the search space using reduced parameters bounds from the best iterations of object similar to the new one, guarantees overall a faster and better optimization with a mean grasping success of 84.3\% vs 78.9\% without meta-learning. Moreover, the increase in the mean expected performance from the optimization with meta-learning is consistent for every object tested, simulated or real (75.9\% vs 68.4\%, 79.4\% vs 77.6\%, 89.4\% vs 65.1\%, 93.1\% vs 91.0\%, 86.7\% vs 86.0\%, 76.9\% vs 76.0\%, 88.9\% vs 88.1\%, and 84.6\% vs 75.6\%), and is stronger for object presenting a higher challenge. When considering only the best run for each object among the 6, the optimization with meta-learning reaches 80.0\%, 85.0\%, 96.1\%, 98.9\%, 90.0\% and 83.3\% and 93.3\% for respectively object A, C1, D, P1, \textit{hammer$\_$j}, \textit{m784} and \textit{cokeSmallGrasp}, which represents a mean score of 89.5\%.\\

One of the assumption in this work was that the default parameters bounds where large enough to include optimized values within the range, that is why the Parameters Bounds module has been designed to only reduced them. However, future work will investigate the possibility of the parameters bounds to also be extended, which can be useful in particular when the manually defined default bounds are too constrained for a specific task.

We aim also to use this developmental learning framework from simulation into a transfer learning setup, where the reduced parameters bounds and the optimized parameters of a simulated object O will be used when optimizing the same object O but with a real robot, as explored for grasping problems recently\cite{breyer2018flexible}. The robot will use its simulated experiences in order to warm-start and simplify the optimization of the bin-picking of the same object when confronted in reality. The use of the simulation applied to transfer learning has the benefit of allowing the robot to always train and learn "mentally" (\textit{i.e.} when one computer is available, and can even "duplicate" itself and run multiple simulation from several computers) even if the physical robot is already used or is costly to run, which is the case usually for industrial robots \textit{in-situ}.

Eventually, this work can be extended toward the developmental embodied aspect of the robotics field, when reduced parameters bounds might potentially be linked to embodied symbols or concept emergence~\cite{taniguchi2016symbol} related to physical properties of the manipulated objects. A possible method to investigate such properties would be to find co-occurrences between sub-set of reduced parameters bounds and human labels or description of the object (\textit{e.g.} "flat", "heavy") or of the manner the task has been achieved (\textit{e.g.} "fast"), in a similar way that was done to discover pronouns~\cite{pointeau2014emergence} or body-parts and basic motor skills~\cite{petit2016hierarchical}. This would allow in return a possible human guidance in an intuitive manner to the robot by constraining the search space based on the label provided by the human operator.

%\section*{References}

%begin{thebibliography}{00}
\balance
\bibliographystyle{IEEEtran}
\bibliography{./refs.bib}
%end{thebibliography}

\end{document}